\documentclass[10pt,twocolumn,letterpaper]{article}

\usepackage{iccv}              % To produce the CAMERA-READY version
\usepackage{graphicx}
\usepackage{subcaption}
\usepackage{amsmath}
\usepackage{arydshln} 
\usepackage{nccmath} 
\usepackage{url}
\usepackage{mathtools}
\usepackage{multirow}
\usepackage{color}
\usepackage{adjustbox}
\usepackage[font=small,skip=2pt]{caption}
\newcommand{\cmmnt}[1]{}

\usepackage{amssymb}
\usepackage{pifont}
\usepackage{newunicodechar}
\newunicodechar{✓}{\ding{51}}
\newunicodechar{✗}{\ding{55}}
\newcommand{\R}{\ensuremath{\mathbb{R}}}
\definecolor{gold}{rgb}{1.0, 0.84, 0.0}
\definecolor{orchidmagenta}{rgb}{0.85, 0.44, 0.84}
\definecolor{iccvblue}{rgb}{0.21,0.49,0.74}
\usepackage[pagebackref,breaklinks,colorlinks,allcolors=iccvblue]{hyperref}

\def\paperID{} %13886% *** Enter the Paper ID here
\def\confName{ICCV}
\def\confYear{2025}

\newcommand*{\email}[1]{\texttt{#1}}

\title{BeatFormer: Efficient motion-robust remote heart rate estimation through unsupervised spectral zoomed attention filters}

\author{
Joaquim Comas and Federico Sukno\\
 \normalsize Department of Information and Communication Technologies, Pompeu Fabra University, Barcelona, Spain\\
\small \email{\{joaquim.comas,federico.sukno\}@upf.edu}\\
}

\begin{document}
\maketitle
\begin{abstract}

Remote photoplethysmography (rPPG) captures cardiac signals from facial videos and is gaining attention for its diverse applications. While deep learning has advanced rPPG estimation, it relies on large, diverse datasets for effective generalization. In contrast, handcrafted methods utilize physiological priors for better generalization in unseen scenarios like motion while maintaining computational efficiency. However, their linear assumptions limit performance in complex conditions, where deep learning provides superior pulsatile information extraction. This highlights the need for hybrid approaches that combine the strengths of both methods. To address this, we present BeatFormer, a lightweight spectral attention model for rPPG estimation, which integrates zoomed orthonormal complex attention and frequency-domain energy measurement, enabling a highly efficient model. Additionally, we introduce Spectral Contrastive Learning (SCL), which allows BeatFormer to be trained without any PPG or HR labels. We validate BeatFormer on the PURE, UBFC-rPPG, and MMPD datasets, demonstrating its robustness and performance, particularly in cross-dataset evaluations under motion scenarios. The code is available at our \href{https://qcomas7.github.io/BeatFormer.github.io/}{project website}.

\end{abstract}
\section{Introduction}
\label{sec:intro}

In recent years, interest in camera-based physiological signal measurement has grown rapidly due to its potential in clinical \cite{huang2021neonatal} and human-computer interaction applications \cite{benezeth2018remote, nowara2020near}. Deep learning has accelerated progress in this field, but data-driven approaches heavily depend on training data, making them sensitive to biases in skin tone, lighting, video compression, and body motion, which impact model robustness and fairness.

While data-driven models outperform handcrafted methods in many cases, the recent MMPD benchmark \cite{tang2023mmpd} showed that traditional approaches like POS \cite{wang2016algorithmic} still achieve superior performance in some motion scenarios by leveraging physiological knowledge. This raises questions about the superiority of computationally expensive models. Data-driven methods often struggle with generalization, particularly on small datasets, whereas handcrafted approaches benefit from physiological priors, enabling efficient and robust rPPG measurement. However, their reliance on linear projections limits their ability to capture non-linear physiological relationships, reducing their effectiveness in real-world conditions.

%To overcome motion bias, recent data-driven approaches have proposed various solutions, including preprocessing steps \cite{maity2022robustppg}, optical flow integration \cite{li2023learning}, and motion-based data augmentation \cite{paruchuri2024motion}. However, some of these methods rely on computationally expensive preprocessing stages or require external data to mitigate motion bias. Regarding traditional approaches, chrominance-based models \cite{de2013robust, wang2016algorithmic}, demonstrated robust pulsatile projections to motion artifacts. To address linearity constraint, the frequency domain was explored \cite{wang2017robust, wang2017color} by considering sub-band frequency RGB decomposition. However, one of the limitations of these frequency domain approaches is their reliance on the Fast Fourier Transform (FFT) due to its frequency trade-off resolution, particularly in short time intervals. 

To mitigate motion bias, recent deep learning-based methods have explored preprocessing techniques \cite{maity2022robustppg}, optical flow integration \cite{li2023learning}, and motion-aware data augmentation \cite{paruchuri2024motion}. However, many of these solutions require expensive preprocessing or external data. Traditional chrominance-based models \cite{de2013robust, wang2016algorithmic} have demonstrated robustness to motion, while improved frequency-domain approaches \cite{wang2017robust, wang2017color} decompose RGB signals into sub-bands for better signal separation. However, these methods typically rely on the Fast Fourier Transform (FFT), which suffers from resolution trade-offs, which do not guarantee motion separation, particularly in short time intervals.

%Recent work \cite{comasdeepzoom2024} introduced an adaptive frequency heart rate estimator based on the Chirp-Z Transform (CZT) \cite{rabiner1969chirp}, a generalization of the Discrete Fourier Transform (DFT). In contrast to the DFT, the CZT enables adjustable resolution and frequency range, allowing targeted analysis within a specific band, such as the HR bandwidth. This flexibility, as demonstrated in the study, improves HR estimation accuracy even with short temporal windows of the rPPG signal. While CZT has shown promise for remote HR estimation, its potential for enhancing rPPG signal recovery remains unexplored.

A recent study \cite{comasdeepzoom2024} introduced an adaptive frequency-domain heart rate estimator using the Chirp-Z Transform (CZT) \cite{rabiner1969chirp}, an extension of the Discrete Fourier Transform (DFT) that allows for adjustable frequency resolution and range. Unlike the DFT, the CZT enables targeted spectral analysis within specific bands, such as the heart rate bandwidth, improving accuracy even with short temporal windows. While CZT has shown promise for remote heart rate estimation, its potential for enhancing rPPG signal recovery remains unexplored.

%Inspired by the works of Yang et al. \cite{yang2020complex} and Wang et al. \cite{wang2017color}, we aim to learn a set of spectral filters that project RGB skin content into a space where pulsatile information is effectively separated from distortion artifacts. To achieve this, we introduce BeatFormer, a lightweight spectral attention model composed of Zoomed Orthonormal Complex Attention (ZOCA) blocks and a spectral zoomed energy density measurement.

Inspired by Yang et al. \cite{yang2020complex} and Wang et al. \cite{wang2017color}, we propose BeatFormer, a lightweight spectral attention model that learns spectral filters to separate pulsatile information from motion variations. BeatFormer consists of Zoomed Orthonormal Complex Attention (ZOCA) blocks and a spectral zoomed energy density measurement, leveraging frequency-domain features for robust rPPG estimation. Instead of relying on unconstrained data-driven weights, it incorporates implicit physiological priors, enhancing robustness against training data noise (e.g., illumination changes and motion) while reducing computational cost and parameter count compared to existing models. Additionally, by integrating explicit priors through unsupervised learning, BeatFormer achieves performance comparable to supervised methods without requiring labeled data. This efficiency in both parameters and training enables strong generalization, even in label-free settings.

%we impose explicit and implicit physiological priors by adopting frequency self-supervised learning to enhance robustness against training data noise (e.g., illumination variations and body motion). Besides these priors help reducing the computational cost as well as the number of parameters compared to existing data-driven models. As a result, BeatFormer is not only highly efficient in terms of parameters but also in its training process, leading to strong generalization even under relatively small training sets. %requiring less training data to achieve strong generalization.

The main contributions of the paper are three-fold: 
\begin{itemize}
    \item We introduce BeatFormer, a lightweight spectral filter transformer for rPPG estimation, integrating zoomed orthonormal complex attention and frequency-domain energy measurement. Our approach combines data-driven modeling with implicit physiological priors, ensuring generalization comparable to traditional methods while remaining computationally efficient.
    
    \item A frequency self-contrastive learning approach based on explicit physiological assumptions is introduced, offering robustness comparable to supervised learning while eliminating the need for labeled data, even under significant motion distortions.

    \item Extensive experiments on publicly available rPPG benchmark datasets highlight the benefits of frequency-domain rPPG estimation, particularly in motion scenarios, using the CZT for its promising properties in rPPG estimation.
    
\end{itemize}

% Contribution + organization paper

%implicit physiological priors
%efficient model in two points of view, in terms of computational cost and parameters as well as in terms of data training need
%
\section{Related work}
\label{sec:related}
%\subsection{Camera-based PPG measurement}

%Since the pioneering works of Takano et al. \cite{takano2007heart} and Verkruysse et al. \cite{verkruysse2008remote}, researchers have explored various techniques for remote heart rate estimation. Traditional methods focus on defining regions of interest and applying signal processing techniques like Blind Source Separation \cite{poh2010non, poh2010advancements} and Normalized Least Mean Squares \cite{li2014remote}, while others leverage skin optical reflection models to reduce motion artifacts \cite{de2013robust, wang2016algorithmic, wang2017color}.

%More recently, deep learning has transformed the field, surpassing classical approaches in accuracy \cite{vspetlik2018visual, yu2019remote, perepelkina2020hearttrack, lu2021dual, comas2024deep}. Some models integrate CNNs with traditional techniques \cite{niu2019rhythmnet, song2021pulsegan}, while others adopt fully data-driven, end-to-end architectures \cite{chen2018deepphys, Yu2019RemotePS}. Transformer-based models \cite{yu2023physformer++, gupta2023radiant, liu2024rppg} have further improved spatiotemporal feature extraction but remain computationally demanding, prompting the development of lightweight alternatives \cite{liu2023efficientphys, comas2022efficient}.

\textbf{Camera-based PPG measurement}. Since the pioneering works of Takano et al. \cite{takano2007heart} and Verkruysse et al. \cite{verkruysse2008remote}, researchers have developed various techniques for remote heart rate estimation. Traditional methods rely on defining regions of interest and applying signal processing techniques like Blind Source Separation \cite{poh2010non, poh2010advancements} and Normalized Least Mean Squares \cite{li2014remote}, while others use skin optical reflection models to reduce motion influence \cite{de2013robust, wang2016algorithmic, wang2017color}.

Deep learning has since transformed the field, surpassing classical methods in accuracy \cite{vspetlik2018visual, yu2019remote, perepelkina2020hearttrack, lu2021dual, comas2024deep}. Some models combine CNNs with traditional techniques \cite{niu2019rhythmnet, song2021pulsegan}, while others adopt end-to-end architectures \cite{chen2018deepphys, Yu2019RemotePS}. Transformer-based models \cite{yu2023physformer++, gupta2023radiant, liu2024rppg} further improve spatiotemporal feature extraction but remain computationally demanding, motivating research on lightweight alternatives \cite{liu2023efficientphys, comas2022efficient}. Beyond supervised learning, researchers are addressing generalization challenges through unsupervised strategies like meta-learning \cite{lee2020meta, liu2021metaphys} and contrastive learning \cite{gideon2021way, sun2022contrast}. Domain adaptation techniques \cite{lu2023neuron, du2023dual, chari2024implicit} and data augmentation methods \cite{paruchuri2024motion, comas2024physflow, hsieh2022augmentation} further mitigate biases related to motion, skin tone, and heart rate distribution.

%\subsection{Motion solutions in rPPG estimation}
\textbf{Motion solutions in rPPG estimation}. Body motion remains a key challenge in rPPG estimation. Early methods such as CHROM \cite{de2013robust} and POS \cite{wang2016algorithmic} projected skin pixels into optimized subspaces to reduce motion noise, while frequency-based approaches \cite{wang2017robust, zhou2020enhancing} decomposed RGB signals for improved robustness. On the other hand, data-driven methods have introduced new strategies, such as optical flow-based motion estimation \cite{li2023learning, wu2023motion} and preprocessing techniques like 3D inverse rendering \cite{maity2022robustppg}, orientation-conditioned facial mapping \cite{cantrill2024orientation}, and motion-transfer augmentation \cite{paruchuri2024motion}. Recently, masked attention mechanisms \cite{zhao2024toward} have been proposed to enhance motion resilience. Despite these advancements, motion mitigation remains an open problem. Many deep learning methods rely on computationally expensive preprocessing steps, usually requiring external components for generalization, while the frequency domain is largely unexplored in data-driven approaches. Consequently, developing efficient and robust motion methods remains an open challenge in rPPG research.

%\subsection{Spectral Attention-based modeling}
\textbf{Spectral Attention-based modeling} The rise of attention mechanisms \cite{vaswani2017attention} has transformed artificial intelligence, impacting various tasks \cite{plizzari2021spatial, liu2022learning, liu2021fuseformer}. While most time series models use temporal attention, some studies explore the spectral domain. Early work \cite{wolter2018complex, trabelsi2018deep} integrated complex values into recurrent networks, achieving state-of-the-art results. Yang et al. \cite{yang2020complex} later introduced the first complex transformer for Automatic Music Transcription. Other studies \cite{qin2021fcanet, lee2021fnet} leveraged DCT and FFT to develop efficient spectral attention models. More recently, Kang et al. \cite{kangintroducing} proposed spectral attention for long-range dependencies in time series forecasting. Despite the strong link between rPPG periodicity and the frequency domain, most rPPG methods \cite{yu2021transrppg, yu2022physformer, zhang2023demodulation, gupta2023radiant} primarily rely on temporal attention, often overlooking spectral information.

%\subsection{Self-contrastive learning}
\textbf{Self-contrastive learning}. Contrastive learning has gained attention for its success in self-supervised representation learning, particularly in computer vision tasks \cite{schroff2015facenet, chen2020simple, qian2021spatiotemporal}. It enhances model training by maximizing intra-class similarity and minimizing inter-class differences. Recently, it has been applied to rPPG signal recovery. Gideon et al. \cite{gideon2021way} pioneered its use for training a saliency sampler to extract rPPG signals. Other works \cite{sun2022contrast, sun2024contrast} employ spatio-temporal samplers for unsupervised and weakly supervised learning. Birla et al. \cite{birla2023alpine} used contrastive learning to capture temporal similarities across multiple ROIs. Recent transformer-based approaches \cite{wang2023transphys, savic2024rs} incorporate contrastive learning using spatiotemporal augmentations or chrominance-based methods like \cite{de2013robust, wang2016algorithmic}. Building on these advancements, we focus our contrastive learning on video transformations that provide meaningful frequency representations, serving as explicit physiological priors.

\begin{figure*}[t]
    \centering
\includegraphics[width=0.89\textwidth]{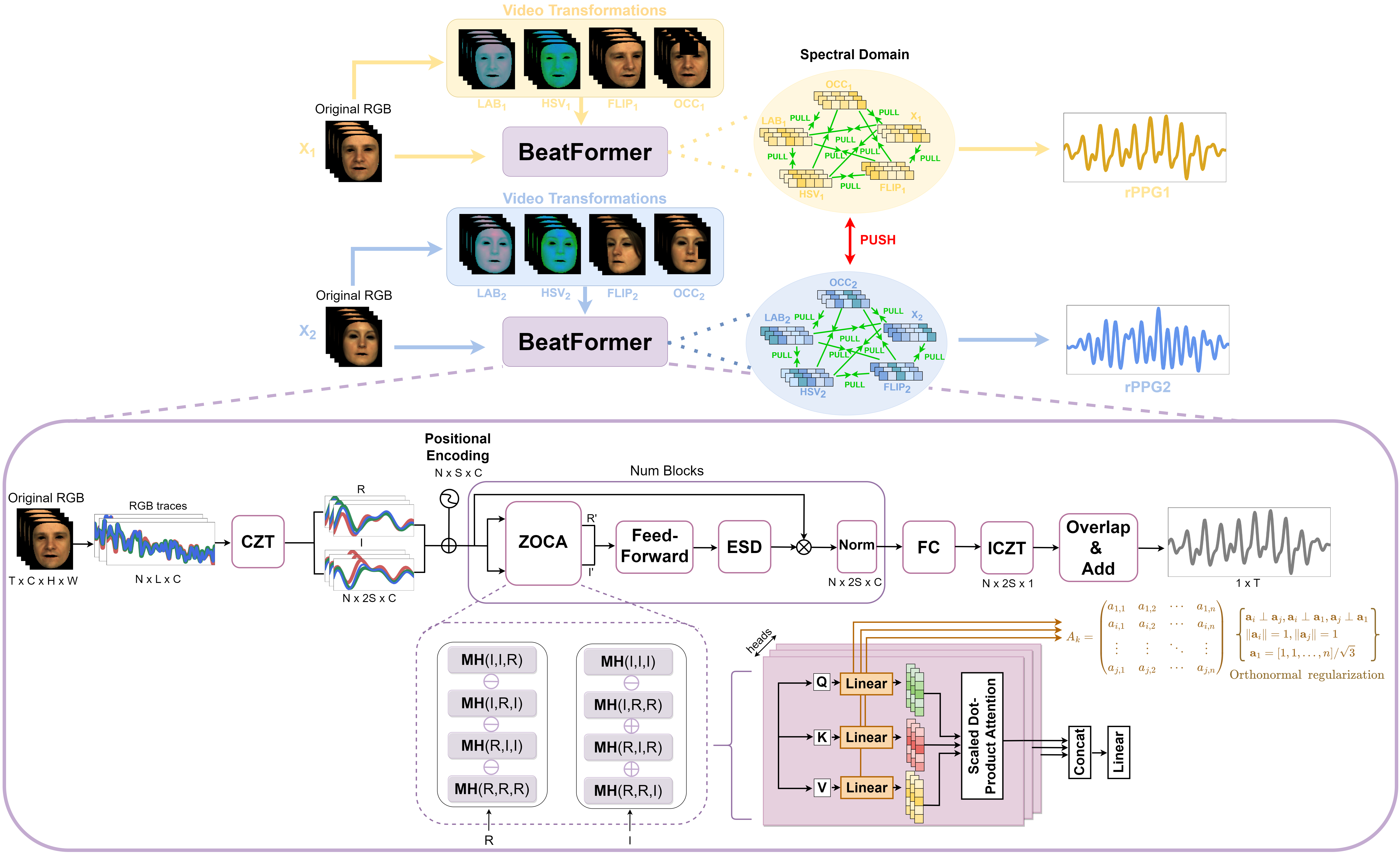}
    \caption{{BeatFormer overall structure. First, RGB traces are segmented with overlap and transformed into the frequency domain using the CZT. The zoomed spectrum is then processed by BeatFormer to filter pulsatile information from distortions, incorporating orthonormal regularization and energy-based weighting. The filtered frequency features are converted back to the temporal domain using the ICZT, followed by an overlap-add operation to reconstruct the rPPG signal. To train BeatFormer, spectral contrastive learning (SCL) is applied, leveraging frequency-domain meaningful transformations to enforce explicit priors during training, enabling label-free training.}} 
    \label{fig:BeatFormer}
\end{figure*}

\section{Methodology}
 In this section, we introduce BeatFormer, depicted in Fig. \ref{fig:BeatFormer}. First, in Subsection \ref{sec:czt} we briefly review the CZT, while subsections \ref{sec:beatformer} and \ref{sec:contrastive} explain the proposed model and its training optimization.

\subsection{Preliminaries: Chirp-Z Transform }
\label{sec:czt}
The CZT, originated in 1969 by Rabiner et al. \cite{rabiner1969chirp}, computes the z-transform of the finite duration signal $x[n]$ along a general spiral contour in the z-plane. Therefore, the CZT is defined  using the following formula:
\begin{equation}
\centering
CZT(x[n]) = \sum_{n=0}^{N-1} x[n] \cdot z_{k}^{-n}
\end{equation} 
Unlike the DFT, which evaluates the Z-transform of $x[n]$  on $N$ equally spaced points on the unit circle in the  z-plane, the CZT is not constrained to operate along the unit circle, evaluating the z-transform along spiral contours described as:
\begin{equation}
z_k = A \cdot W^{-k},  \quad  \text{\footnotesize{$k= 0,1, ..., M-1$}}
\end{equation}
\noindent where A is the complex starting point, W is a complex scalar describing the complex ratio between points on the contour, and M is the length of the transform. In addition, the CZT can also be expressed with the following matrix expression: 
\\\\
\begin{equation}
\renewcommand{\arraystretch}{1}
\setlength{\arraycolsep}{1.2pt}
\resizebox{0.91\linewidth}{!}{
    $\smash{\underbrace{\begin{bmatrix}
        X_0 \\ X_1 \\ X_2 \\ \vdots \\ X_n
    \end{bmatrix}}_{X}}
    = \smash{\underbrace{
    \begin{bmatrix}
        1 & 1 & 1 & \cdots & 1 \\
        1 & W^1 & W^2 & \cdots & W^{(n-1)} \\
        1 & W^2 & W^4  & \cdots & W^{2(n-1)} \\
        \vdots & \vdots & \vdots & \ddots & \vdots \\
        1 & W^{(m-1)} & W^{2(m-1)} & \cdots & W^{(m-1)(n-1)}
    \end{bmatrix}}_{W}}
    \smash{\underbrace{
    \begin{bmatrix}
        A^{-0} & 0 & 0 & \cdots & 0 \\
        0 & A^{-1} & 0 & \cdots & 0 \\
        0 & 0 & A^{-2}  & \cdots & 0 \\
        \vdots & \vdots & \vdots & \ddots & \vdots \\
        0 & 0 & 0 & \cdots & A^{-n}
    \end{bmatrix}}_{A}}
    \smash{\underbrace{\begin{bmatrix}
        x_0 \\ x_1 \\ x_2 \\ \vdots \\ x_n
    \end{bmatrix}}_{x}}$
}
\label{eq:matrixczt}
\end{equation}
\noindent where A is a $N$ by $N$ diagonal matrix and W is an $M$ by $N$ Vandermonde matrix. 

In this work, we explore the CZT as an alternative to FFT, enabling narrow temporal window analysis without losing frequency resolution due to its spectral zoom property. To focus on the relevant rPPG frequency spectrum, we restrict the CZT to the region of the unit circle, corresponding to the HR bandwidth. Specifically, we define a zoom region starting at $A$ and ending at $(M-1)W$. Based on existing literature and common rPPG datasets, we constrain the HR band to 0.66-2.5 Hz, corresponding to 40–150 beats per minute (BPM). Furthermore, CZT offers configurable bin density, allowing for enhanced frequency resolution within a specified spectrum region. Following \cite{comasdeepzoom2024}, we set the CZT size $M$ equal to the input size $N$. This means, at the typical sampling rate of 30 frames per second, CZT has approximately 13 times higher frequency resolution than FFT in HR bandwidth, i.e. the same number of bins is used to cover $1.84$ Hz instead of 30 Hz.

\subsection{BeatFormer}
\label{sec:beatformer}

\subsubsection{Preprocessing}
Given $C(t) \in \R^{T\times 3}$, as the spatially averaged RGB traces \cite{poh2010advancements, wang2016algorithmic} from the skin facial region captured by the camera for a sequence of $T$ frames, we first temporally normalized $C(t)$ to remove its dependency from DC components, typically corresponding to illumination level: 

\begin{equation}
    C'(t) = \dfrac{C(t)}{\mu(C(t))}-1
\end{equation}
where $C'(t)$ represents the zero mean signal for each  RGB channel and $\mu$ denotes the signal's average value. The temporal normalized signal $C'(t) \in \R^{T\times 3}$ is then divided into overlapping windows of size $L$, yielding $X\in \R^{N\times L \times 3}$, where $N=T-L+1$ is the number of segments.

\subsubsection{Zoomed orthonormal complex attention}
Then, we transform the temporal windowed signal to the frequency domain using CZT: 
\begin{equation}
    F = CZT(X(t))
\end{equation}
where $X(t)$ is the windowed normalized signal, and $F \in \mathbb{C}^{N\times 2M \times C}$ is the zoomed RGB frequency spectrum within the HR bandwidth. Here, $N$ is the number of segments, $M$ represents the number of subbands of size $L$ (as defined in subsection \ref{sec:czt}) covering the 0.66–2.5 Hz range, and $C$ the number of channels. To mitigate training instabilities with complex numbers, in the backpropagation process \cite{tan2022real}, we decompose the spectrum as $F = R + I$, where $R$ and $I$ are the real and imaginary parts, following Euler’s formula.

Before feeding the spectral component $F$ into the BeatFormer, we incorporate a trainable positional encoding to preserve the subband ordering of the spectrum. To formulate our zoomed spectral orthonormal filtering we incorporate the complex attention from \cite{yang2020complex}. Given our RGB frequency complex input $F = R + I$, we can express the queries, keys and values as $Q = FW_Q$, $K = FW_K$ and $V = FW_V$, respectively, where $W_i \in Q, K, V$ are the learnable weights. Therefore, we define the $QK^TV$ dot product as follows:
\vspace{-1pt}
\begin{equation}
\resizebox{0.905\linewidth}{!}{$
\begin{aligned}
    QK^TV&= (FW_Q) (FW_K) (FW_V) \\
    &=(RW_{Q} + IW_{Q})(RW_{K} + IW_{K})(RW_{V} + IW_{V}) \\ 
\end{aligned}
$}
\end{equation}
\vspace{-1pt}
Developing the above complex matrix multiplication, we obtain four complex attention blocks for real and imaginary parts respectively. For each block, the scaled complex dot-product attention is expressed as: 
\vspace{-1pt}
\begin{equation}
    Att(Q,K,V) = \operatorname{Min-max-Norm}(\frac{QK^T}{\sqrt{d_k}})V
\end{equation}
\vspace{-1pt}
where the dot product between its query and all the keys is calculated for each given frequency subband. The resulting value is scaled by the square root of $d_q$ (the frequency features dimensionality), followed by a min-max normalization operation, instead of the common softmax operation, replaced for computational stability in the presence of complex numbers \cite{yang2020complex}. The obtained scores associated with each frequency channel transform them into a weighted sum of the features from all the signal channels.

The multi-head complex attention output is formed by concatenating the outputs of all attention heads, followed by a linear projection using the weight matrix $W^O \in \R^{C\times C}$.
\vspace{-1pt}
\begin{equation}
MH(Q,K,V) = Concat(Att_0, Att_1,...,Att_h)W^O    
\end{equation}
\vspace{-1pt}
Here, $h$ is the total number of heads. Finally, we can define the complex attention ($CA$) as: 
\begin{equation}
\resizebox{0.90\linewidth}{!}{$
\begin{split}
     CA(F) = (MH(R,R,R)-MH(R,I,I)-MH(I,R,I)\\
     -MH(I,I,R)) +(MH(R,R,I)+MH(R,I,R)\\
     +MH(I,R,R)-MH(I,I,I))
\end{split}
$}
\end{equation}

With this configuration, a complex transformer can be designed and directly applied for rPPG estimation. However, in this work, we propose constraining the learning of these complex attention blocks for two key reasons. First, it enhances generalization while preventing overfitting to the training data. Second, it reduces the number of parameters, resulting in a more efficient framework. To achieve this, after defining the complex attention mechanism, we regularize the learned attention weights from $CA$ enforcing implicit physiological priors. Inspired by \cite{wang2016algorithmic, wang2017color}, we constrain the first row of each attention weight to the unit-length vector $[1,1,1]/\sqrt3$, eliminating intensity variations in this direction. Additionally, the subsequent attention rows are learned during training, but forcing an orthonormal relationship between them. To constraint the training process we introduce a regularization orthonormal loss defined as:

%\begin{equation}
%\mathcal{L}_{or} = \frac{1}{N} \sum_{k=1}^{N} \Bigg[ 
%\underbrace{\sum_{i<j} \left( \mathbf{a}_{i}^{(k)} \cdot \mathbf{a}_j^{(k)} \right)^2}_{\mathclap{\text{Orthogonality}}} + 
%\underbrace{\sum_{n} \left( \|\mathbf{a}_n^{(k)}\| - 1 \right)^2}_{\mathclap{\text{Unit norm}}} 
%\Bigg]
%\label{or_loss}
%\end{equation}

\begin{equation}
\small 
\mathcal{L}_{or} = \frac{1}{N} \sum_{k=1}^{N} \bigg[ 
\smash{\underbrace{\sum_{i<j} \left( \mathbf{a}_{i}^{(k)} \cdot \mathbf{a}_j^{(k)} \right)^2}_{\mathclap{\text{\scriptsize Orthogonality}}}} + 
\smash{\underbrace{\sum_{n} \left( \|\mathbf{a}_n^{(k)}\| - 1 \right)^2}_{\mathclap{\text{\scriptsize Unit norm}}}} 
\bigg]
\label{or_loss}
\vspace{+0.5cm}
\end{equation}

%$N$ is the total number of attention matrices, and \textbf{}

Here, $\mathbf{a}_i^{(k)}, \mathbf{a}_j^{(k)}$ are the row vectors of the \(k\)-th $Att$ weights matrix. In our configuration, they denote the second and third rows of attention matrices, respectively. %The first loss term enforces orthogonality between attention rows, while the second ensures orthonormality, where \( \|\cdot\| \) is the L2-norm. 

\subsubsection{Energy-measurement feed-forward}

Unlike the standard transformer feed-forward mechanism, we incorporate energy contribution measurement between frequency subbands in the zoomed HR bandwidth to filter out distortions from the pulsatile information. %to filter pulsatile information from distortions.

After applying zoomed orthonormal complex attention (ZOCA),  its output $Z$ is passed through two multilayer perceptron layers with GELU activation, yielding $Z'\in \mathbb{C}^{N\times 2M \times 1}$. 
Since there are N segments, 2M real and imaginary subbands, %Here, $N$ is the number of segments, $2M$ represents subbands for real and imaginary components, 
and the last dimension corresponds to the fused frequency channels. Then, we measure the contribution of each learned subband weighting by computing the energy contribution using the energy spectral density (ESD):  
\begin{equation}
S = \frac{|Z'(f)|^2}{|F(f)|^2} = \frac{Z'(f) \cdot Z'^*(f)}{F(f) \cdot F^*(f)}
\end{equation}
where \( Z'^*(f) \) and \( F^*(f) \) are the complex conjugates of \( Z'(f) \), the ZOCA output, and \( F(f) \), the RGB frequency input, respectively. The idea behind the energy measurement relies on the assumption that pulsatile and motion signals exhibit different relative amplitudes across the RGB channels \cite{wang2017color}. Then, instead of a residual connection, we multiply frequency weights by the input frequencies to learn pulse signal filtering. The channel dimension of $S$ is expanded to match the RGB channels, and energy contribution filtering is normalized to prevent gradient explosion during recursive attention processing: 
\begin{equation}
F' = Norm( F \cdot S) 
\end{equation}
To reconstruct the pulse signal, a multilayer perceptron fuses channel information into a single output before applying the inverse Z transform (ICZT) to the filtered spectrum, yielding a one-dimensional windowed signal $P' \in N\times L\times 1$. The final pulse signal $P$  is obtained by merging overlapping segments via an overlap-add operation \cite{de2013robust}. 

Incorporating hand-crafted techniques such as spatially averaging RGB traces, energy measurement contribution, orthonormality regularization and the overlap-addition operation helps considerably reduce the number of parameters and prevents overfitting the learnable parameters to the training data, making an efficient and robust rPPG solution.

\subsection{Spectral contrastive learning}
\label{sec:contrastive}

Our second contribution enables BeatFormer to be trained in an unsupervised manner, achieving almost the same performance as in the supervised case, but without any PPG or HR information from the training videos. By eliminating the need for labeled data, BeatFormer reduces reliance on dataset biases (e.g., lack of motion variations, corrupted PPG labels) and ensures robust generalization. %, enhancing training efficiency without relying on labeled data, while achieving results comparable to supervised training. 
 %, improving performance even with a reduced amount of unlabeled data.

To achieve this unsupervised training, we propose spectral contrastive learning (SCL), which applies video transformations in the frequency domain as explicit priors. This approach ensures that the model captures relevant patterns in the data without requiring labeled examples. Since BeatFormer operates in this domain, we design transformations based on physiological assumptions (e.g., characteristics of human motion or heart rate), which influence the real and imaginary components (magnitude and phase) of the extracted RGB traces. In contrast to temporal contrastive learning, this method enhances motion and distortion separation by incorporating additional information to spectral magnitude, by including the phase information, which leads to a more meaningful data representation.

To generate the proposed video transformations, we follow three assumptions:

\begin{itemize}
    \item Facial video pulse information remains consistent under different illumination conditions or color representations.
    
    \item rPPG signal exhibits quasi-periodic behavior, whereas body motion follows a more chaotic periodicity.

    \item The cardiac information can be recovered in motion scenarios as long as a sufficient skin region is visible.
\end{itemize}

\begin{figure}[t]
    \centering
\includegraphics[width=0.40\textwidth]{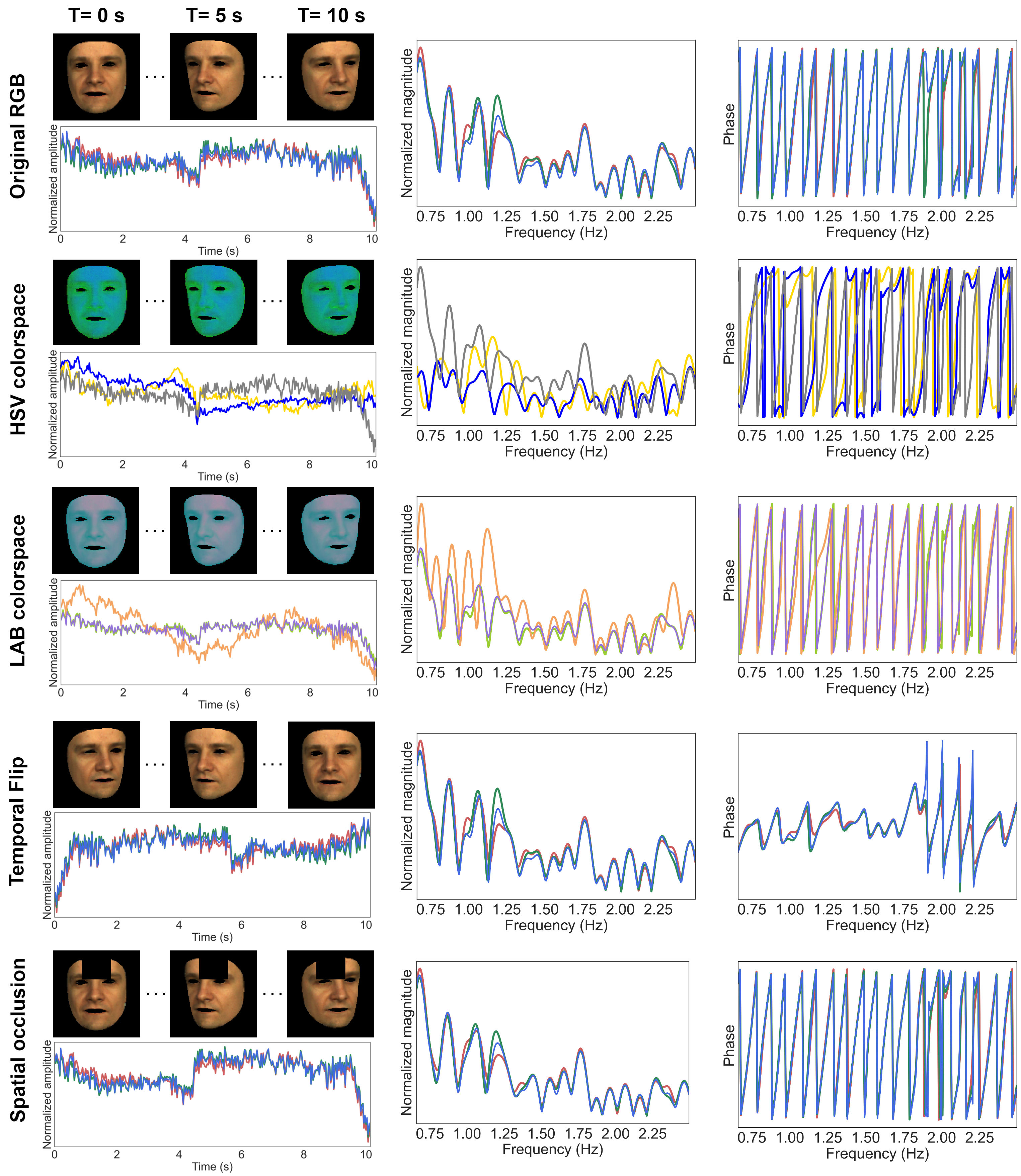}
    \caption{SCL Training Video Transformations (10-sec example): Top to bottom: Original RGB, HSV, and LAB color spaces, temporal flipping, and spatial occlusion. Left: Video transformation and skin-averaged trace evolution. Right: Magnitude and phase frequency representation.} 
    \label{fig:transformations}
     
\end{figure}

%Based on these assumptions, we apply four video transformations (depicted in Fig. \ref{fig:transformations}): HSV and LAB color space conversions, temporal flipping, and random spatial occlusion. HSV and LAB are chosen for their ability to preserve chrominance while separating luminance, improving robustness to illumination changes \cite{dasari2021evaluation} and motion artifacts \cite{yang2016motion}. Additionally, the Hue and Luminance channels have been shown to capture alternative pulsatile information \cite{tsouri2015benefits}. As seen in Fig. \ref{fig:transformations}, while temporal traces and frequency characteristics vary across color spaces, pulsatile information remains consistent (e.g. some channels exhibit similar frequency magnitude behavior). Thus, this diverse pulse representation strengthens the robustness of the model to varying illumination conditions.
%facilitating the separation of pulsatile signals from luminance distortions.

Based on these assumptions, we apply four video transformations (shown in Fig. \ref{fig:transformations}): HSV and LAB color space conversions, temporal flipping, and random spatial occlusion. HSV and LAB are selected for their ability to preserve chrominance while separating luminance, enhancing robustness to illumination changes \cite{dasari2021evaluation} and body motion \cite{yang2016motion}. Besides, Hue and Luminance channels have been shown to capture alternative pulsatile content \cite{tsouri2015benefits}. As shown in Fig. \ref{fig:transformations}, while temporal traces and frequency characteristics vary across color spaces, pulsatile information remains consistent (e.g. some channels exhibit similar frequency magnitude behavior). This diverse pulse representation improves the model's robustness to varying illumination conditions.

For the second assumption, temporal flipping is used since it preserves magnitude information while altering phase information. This property enhances the disentanglement of pulse signals from motion variations in the phase domain, improving robustness even in scenarios with challenging movements throughout a video sequence. Finally, using spatially averaged RGB frames, we limit parameter scalability, enhancing the model efficiency, but also improving robustness against local motions in PPG extraction. To simulate artifacts caused by head movements, we randomly occlude parts of the skin region, emulating temporary occlusions due to motion. As shown in Fig. \ref{fig:transformations}, the temporal and frequency behaviors of the original and occluded examples remain largely consistent, with slight variations such as in those displayed near 1.3 Hz in magnitude or 1.9 Hz in the phase domain. In Section \ref{sec:experiments}, we further analyze the impact of each transformation using SCL training. %across different cross-dataset evaluation scenarios.

To learn from the data itself, without using any labels, we exploit both intra and inter-data relationships. Intra-data learning treats video transformations of the same sample as positive pairs, while inter-data learning considers different samples with their respective transformations as negative pairs in a contrastive manner, as shown in Fig. \ref{fig:BeatFormer}. During BeatFormer training, the rPPG signal's power spectral density is computed for each intra- and inter-sample, adopting the squared Earth Mover’s Distance (EMD) loss \cite{hou2017squared}, used to assess pairwise similarity. Unlike categorical cross-entropy, EMD accounts for inter-class relationships in HR distributions by measuring the minimum cost required to transform one distribution into another, formulated as:
\vspace{-3pt}
\begin{equation}
    \mathcal{L}_{EMD} = \frac{1}{N} \sum_{i=1}^{N} (CDF_{i}(p) - CDF_{i}(t))^2
    \label{emd_loss} \vspace{-3pt}
\end{equation}
\noindent Here, $CDF(\cdot)$ denotes the cumulative density function, while $p$ and $t$ represent two compared distributions of size $N$ (batch dimensionality). In our framework,  $CDF(p)$ and $CDF(t)$ refer to the frequency density functions of the predicted rPPG signals for the two compared pairs.

Thus, our SCL loss is formulated as a triplet loss margin using EMD to enforce similarity between video transformations of the same sample while separating different samples. For intra-sample dissimilarity, we compute:
\vspace{-3pt}
\begin{equation}
    \mathcal{L}_{pos} = \sum_{n=1}^{N} \sum_{i < j} \text{EMD}(p_{n,i}, p_{n,j})
    \vspace{-3pt}
\end{equation}
\noindent where $N$ corresponds to the batch dimensionality, $p_n$ denotes a particular sample, and $(i<j)$ represents unique video transformation pairs within the same sample. On the other hand, the inter-sample dissimilarity is defined as:
\vspace{-3pt}
\begin{equation}
    \mathcal{L}_{neg} = \sum_{x < y}\sum_{i=1}^{V} \sum_{j=1}^{V} \text{EMD}(p_{x,i}, p_{y,j})
    \vspace{-3pt}
\end{equation}
Here, $x$ and $y$ are unique sample pairs, and $V$ is the total number of video transformations. The final SCL loss is formulated as a hinge loss:
\vspace{-3pt}
\begin{equation}
\mathcal{L}_{SCL} = \frac{1}{N} \max\left(0, \mathcal{L}_{pos} - \mathcal{L}_{neg} + \gamma 
\right) \vspace{-3pt}
\end{equation}
\noindent where $\gamma$ is the margin parameter ($\gamma \geq 0$), ensuring that positive pairs remain more similar than negative ones while preventing trivial solutions.
Finally, the total loss combines SCL and the orthonormal regularization:
\vspace{-3pt}
\begin{equation}
\label{combined_eq}
    \mathcal{L}_{total} =  \mathcal{L}_{SCL} + \alpha \cdot \mathcal{L}_{or}
    \vspace{-3pt}
\end{equation}
where $\alpha$ is a balancing parameter. Based on preliminary experiments, we empirically set $\alpha = 0.5$ and $\gamma=1$.

\vspace{-2pt}
\section{Experiments}
\label{sec:experiments}
This section presents the experimental setup, followed by the results, including cross-dataset evaluations and an ablation study of the proposed model. Finally, we will provide a discussion analyzing qualitative results.

\begin{table*}[t]
\centering
\caption{Pulse rate cross-dataset results trained on PURE and UBFC-rPPG and tested on whole MMPD dataset (in BPMs).}
\renewcommand{\arraystretch}{1.015}
\adjustbox{width=0.7\textwidth}{
\begin{tabular}{c| c c c c | c c c c}
\hline
 \multirow{2}{2cm}{\centering Method}  & \multicolumn{4}{c}{PURE $\rightarrow$ MMPD} & \multicolumn{4}{|c}{UBFC-rPPG $\rightarrow$ MMPD} \\ \cline{2-9}
 
 & MAE $\downarrow$ & RMSE $\downarrow$ & MAPE $\downarrow$ & $\rho\uparrow$ & MAE $\downarrow$ & RMSE $\downarrow$ & MAPE $\downarrow$ &$\rho\uparrow$ \\ \hline
 ICA \cite{poh2010non} & 18.57 & 24.28 & 20.85 & 0.00 & 18.57 & 24.28 & 20.85 & 0.00  \\ 
 CHROM \cite{de2013robust} & 13.63 & 18.75 & 15.96 & 0.08 & 13.63 & 18.75 & 15.96 & 0.08  \\ 
 POS \cite{wang2016algorithmic} & 12.34 & 17.70 & 14.43 & 0.17 & 12.34 & 17.70 & 14.43 & 0.17 \\ 
 \hline
 TS-CAN \cite{liu2020multi} & 13.94 & 21.61 & 15.14 & 0.20 & 14.01 & 21.04 & 15.48& 0.24 \\ 
 PhysNet \cite{Yu2019RemotePS} &13.22 & 19.61 & 14.73 & 0.23 & 10.24 & 16.54 & 12.46 & 0.29 \\ 
 DeepPhys \cite{chen2018deepphys}&16.92 & 24.61 & 18.54 & 0.05& 17.50 & 25.00 & 19.27	& 0.05 \\ 
 EfficientPhys \cite{liu2023efficientphys}& 14.03 & 21.62 & 15.32 & 0.17& 13.78 & 22.25 & 15.15 &0.09 \\
 PhysFormer \cite{yu2022physformer}& 14.57 & 20.71 & 16.73 & 0.15 & 12.10 & 17.79 & 15.41 &0.17 \\
 SpikingPhys \cite{liu2025spiking} & 14.57 & - & 16.55 & 0.14 & 14.15 & - & 16.22 & 0.15 \\
PhysNet-UV \cite{cantrill2024orientation} & - & - & - & - & 12.18 & 19.84 & - & 0.29\\
 PhysMamba \cite{luo2024physmamba} & 10.31 & 16.02 & - & 0.34 & 11.96 & 17.69 & - & 0.29 \\
 RhythmFormer \cite{zou2024rhythmformer} & 8.98 & \textbf{14.85} & 11.11 & \textbf{0.51} & 9.08 & \textbf{15.07} & 11.17 &\textbf{0.53} \\
\hline 
\textbf{BeatFormer-SCL (ours)} & 9.14 & 15.13 & 10.78 & 0.40 & 9.25 & 15.39 & 10.93 & 0.36 \\ 
\textbf{BeatFormer-SL (ours)} & \textbf{8.85} & 15.04 & \textbf{10.54} & 0.39 & \textbf{8.98} & 15.16 &\textbf{10.70} & 0.39 \\ \hline 
\end{tabular}
}
\label{tab:cross_SOTA}
\end{table*}

\subsection{Experimental setup} 

\textbf{Data and evaluation protocol}. The proposed model is evaluated on three publicly available datasets: PURE \cite{stricker2014non}, UBFC-rPPG \cite{bobbia2019unsupervised}, and MMPD \cite{tang2023mmpd}, detailed in the supplementary material. Performance is assessed using standard metrics \cite{li2014remote, liu2023rppg}, including mean absolute HR error (MAE), root mean squared HR error (RMSE), mean absolute percentage error (MAPE), and Pearson’s correlation ($\rho$).

%To evaluate the HR estimation performance of the proposed model, all the experiments are conducted on three publicly available datasets: PURE \cite{stricker2014non}, UBFC-rPPG \cite{bobbia2019unsupervised} and MMPD \cite{tang2023mmpd}, which are detailed in our supplementary material. In terms of metric evaluation, we adopt the same metrics used in the literature, such as the mean absolute HR error (MAE), the root mean squared HR error (RMSE), Mean Absolute Percentage Error (MAPE), and Pearson’s correlation coefficients R \cite{li2014remote, liu2021metaphys, liu2023efficientphys}, also defined in the supplementary material. 
%To calculate the performance metrics, the predicted rPPG signal is detrended \cite{tarvainen2002advanced} and filtered using a Butterworth filter with cutoff frequencies of 0.66-2.5 Hz while heart rate is calculated using CZT \cite{comas2024deep}. Our experiments are performed using video-level evaluation with sequences of 300 frames average with 10 frames overlap to compute HR estimation for all reported metrics. For intra-dataset experiments, we follow the protocol outlined in \cite{vspetlik2018visual, lu2021dual}, for the PURE dataset, and we follow the protocol of \cite{lu2021dual} for UBFC-rPPG. For cross-dataset experiments, we follow the same protocols as rPPG-Toolbox \cite{liu2024rppg}. The training datasets are divided into training ($80\%$) and validation ($20\%$). We conducted the experiments by training on the PURE and UBFC-rPPG datasets and evaluating on the MMPD dataset.
The predicted rPPG signal is detrended \cite{tarvainen2002advanced} and filtered with a Butterworth filter (0.66–2.5 Hz), while heart rate is estimated using CZT \cite{comas2024deep}. We perform video-level evaluation with 300-frame sequences and a 10-frame overlap. Cross-dataset experiments follow the protocols of the rPPG-Toolbox \cite{liu2023rppg}. Training datasets are split into 80\% training and 20\% validation; evaluation is done on the MMPD. %trained on PURE and UBFC-rPPG

\subsection{Experimental results}
\begin{figure}[b]
    \centering
\includegraphics[width=0.45\textwidth]{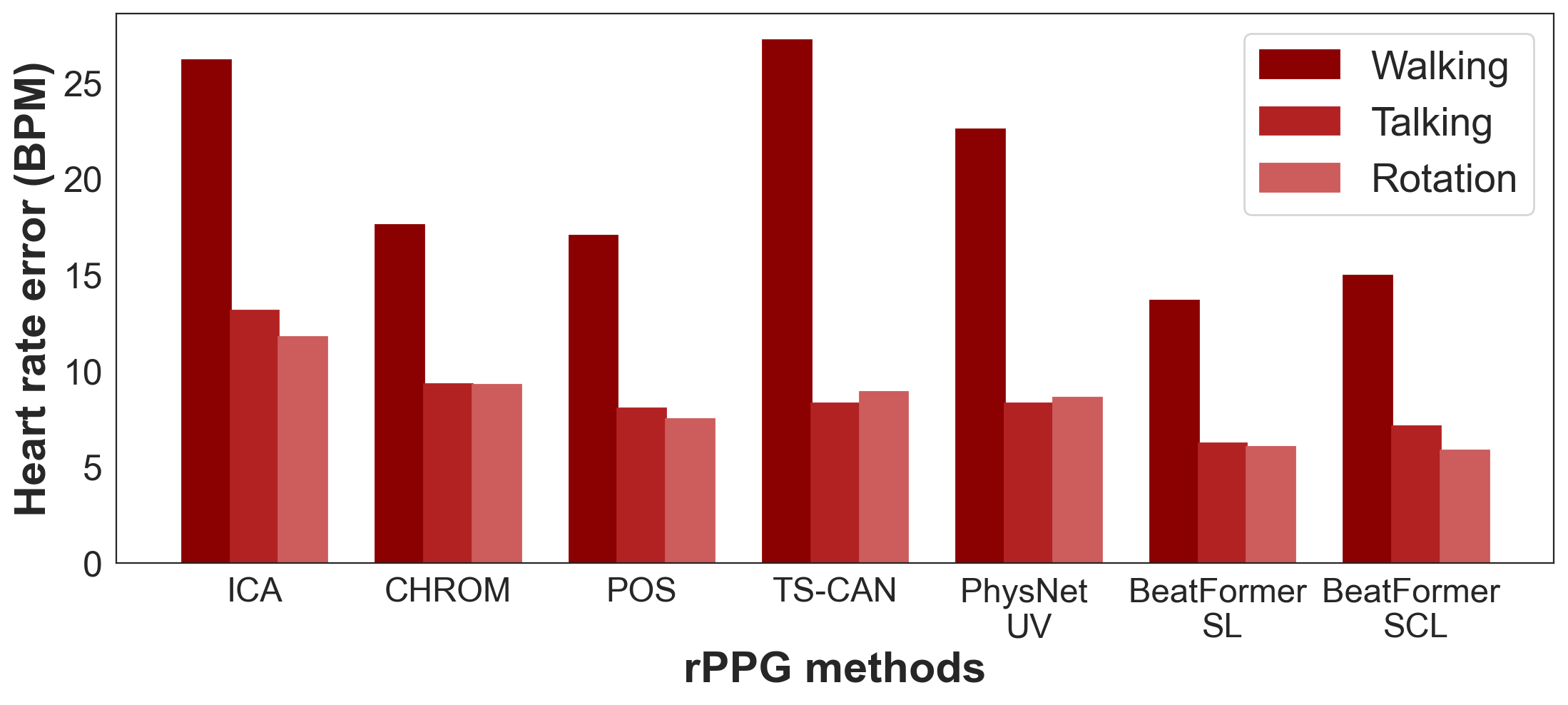}
    \caption{Performance comparison on MMPD motion scenarios.
    %MAE performance comparison in cross dataset evaluation on MMPD motion scenarios. %For DL methods PURE dataset is used as training dataset.
    } 
    \label{fig:transformations_ex}
\end{figure}

\textbf{Implementation details}. In all our experiments, we utilize the Mediapipe Face Mesh model \cite{lugaresi2019mediapipe} to focus our analysis only on facial skin pixels. After masking the facial video, each frame is resized to $96\times 96$ pixels. The PPG ground truth is pre-processed following \cite{dall2020prediction} to denoise the raw signal. We use PyTorch 2.2.2 \cite{paszke2019pytorch} and train on a single NVIDIA RTX3060 using a batch size equals to 2 and sequences of 300 frames without overlap and AdamW optimizer with a Cosine Annealing scheduler \cite{loshchilov2022sgdr} using a maximum learning rate of 5e-4. The proposed model is trained for 20 epochs, for PURE and UBFC-rPPG, with a fixed random seed, using the proposed loss function from Eq. \ref{combined_eq}.

\subsubsection{Cross-dataset evaluation}

Table \ref{tab:cross_SOTA} shows the MMPD cross-dataset results of the Beatformer using its supervised (SL) and unsupervised (SCL) versions compared to the existing rPPG methods, training with reduced datasets like PURE and UBFC-rPPG. For this comparison, handcrafted and data-driven approaches are considered. As expected, traditional methods like POS perform better than costly data-driven models trained on PURE or UBFC-rPPG, due to its greater robustness to the challenging motion videos of the MMPD dataset. Only recent works like PhysNet-UV, PhysMamba and RhythmFormer achieve better results training with these reduced datasets. Regarding BeatFormer, we observe that the supervised version achieves the state-of-the-art similar to RhythmFormer, surpassing MAE and MAPE for both PURE and UBFC datasets. On the other hand, BeatFormer-SCL obtains competitive performance, almost as good as RythmFormer and our supervised version, but without requiring any labels in terms of PPG or HR of the input videos.    

Figure \ref{fig:transformations_ex} shows the impact of motion on rPPG methods across three challenging motion splits of the MMPD dataset, comparing the proposed method to three handcrafted and two data-driven approaches. The walking scenario is the most challenging for all approaches, while rotation and talking yield similar results. Handcrafted methods like CHROM and POS outperform deep learning models (TS-CAN and PhysNet-UV) in all motion scenarios, indicating poor generalization of data-driven methods in complex, unseen conditions very different from the training set. In contrast, BeatFormer (both versions) achieves significantly lower errors across all splits, demonstrating superior robustness to motion. The supervised model performs better in walking and talking, while the unsupervised version slightly outperforms in rotation.

%In contrast, BeatFormer (in both versions) achieves significantly lower error across all splits, showing superior robustness to motion artifacts. Comparing both versions, the supervised model performs better in walking and talking, while the unsupervised version slightly outperforms in rotation.

\begin{table}[t]
\centering
\renewcommand{\arraystretch}{1.1}
\caption{Computational cost comparison.}
\adjustbox{width=0.34\textwidth}{
\begin{tabular}{ccc}
\hline
Method & Params(K) & MACs(M)  \\
\hline
DeepPhys \cite{chen2018deepphys}  &  1980  & 744.45 \\ 
PhysNet  \cite{Yu2019RemotePS} & 768 & 438.24 \\ 
TS-CAN \cite{liu2020multi} & 1980  & 744.45  \\ 
PhysFormer \cite{yu2022physformer} & 7380  & 316.29  \\
RhythmFormer \cite{zou2024rhythmformer} & 3250 &  240.55\\
\hline
\textbf{BeatFormer (ours)} & \textbf{14.86} & \textbf{181.73}\\
\hline  
\end{tabular}
}
\label{table:performance}
\end{table}
\vspace{-2pt}
\subsubsection{Computational cost evaluation}
Table \ref{table:performance} shows the computational cost of several state-of-the-art rPPG approaches. For the comparison, we follow the protocol of \cite{zou2024rhythmformer} and calculate the BeatFormer cost through a Flops counting tool \footnote{https://github.com/sovrasov/flops-counter.pytorch}. The results highlight a significant reduction in computation, as our lightweight model achieves comparable or superior performance with fewer than 15k parameters and 181.73 MACs. By integrating hand-crafted solutions into our data-driven model, specifically, spatially averaged RGB pixels and physiological constraints, BeatFormer avoids the high computational load of data-driven methods that focus on pixel frame level. This demonstrates that fully unconstrained large-scale learning can lead to less efficient and effective models.

\subsection{Ablation Study}
This subsection presents key ablation results for BeatFormer, trained on PURE and tested on MMPD.

\noindent \textbf{Impact of the training loss function.} Table \ref{table:loss_ablation} presents the loss performance ablation study. The experiment uses a temporal L2-loss between the predicted and original PPG signals, the EMD frequential loss (Eq.\ref{emd_loss}), the proposed supervised loss (combining Eq.\ref{emd_loss} and \ref{or_loss}), and the SCL unsupervised loss (Eq.\ref{combined_eq}). The results show that BeatFormer can be optimized with different loss functions to achieve competitive performance. However, in supervised training, the inclusion of orthonormal regularization with the frequential loss outperforms the temporal loss, showing better generalization. Although the proposed SCL loss does not improve MAE, RMSE, or MAPE, it achieves the best Pearson's correlation, with all perfomance metrics remaining close, and without requiring labeled data.

\begin{figure}[t]
    \centering
\includegraphics[width=0.43\textwidth]{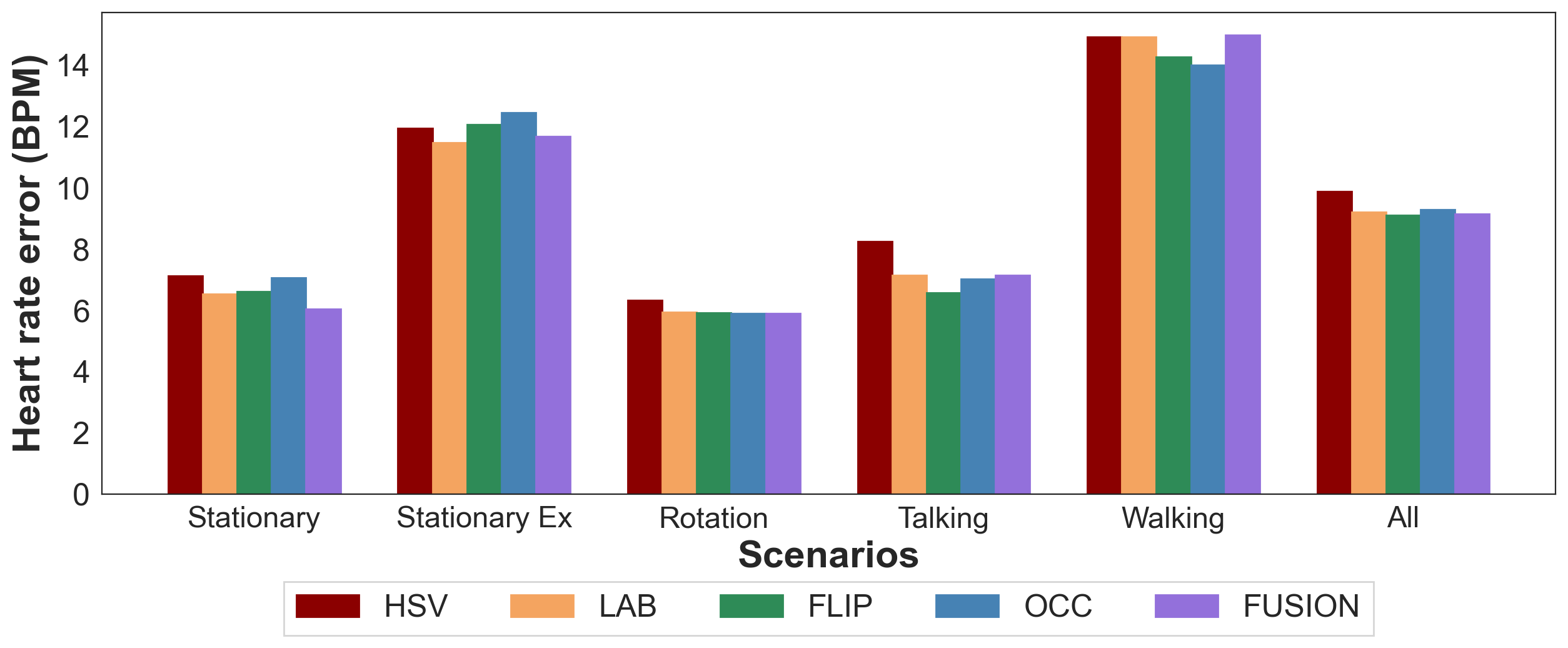}
    \caption{SCL video transformations impact in MMPD splits.  
    } 
    \label{fig:transformation_impact}
\end{figure}
\begin{table}[t]
\centering
\begin{minipage}{0.26\textwidth}
\centering
\renewcommand{\arraystretch}{1.15}
\adjustbox{width=\textwidth}{
  \begin{tabular}{c|c c  c}
    \hline
    Loss function &  MAE $\downarrow$ & RMSE $\downarrow$ & $\rho\uparrow$\\ 
    \hline
    $\mathcal{L}_\mathrm{{MSE}}$ & 8.93 & 15.13  &0.38 \\
    $\mathcal{L}_\mathrm{{EMD}}$ &  8.92 & 15.26 & 0.37 \\ 
    $\mathcal{L}_{EMD} + \mathcal{L}_{OR}$ &  \textbf{8.85} & \textbf{15.04} & 0.39 \\ 
    \hline
    $\mathcal{L}_{SCL} + \mathcal{L}_{OR}$ & 9.14 & 15.13 & \textbf{0.40}\\ 
    \hline
    
  \end{tabular}
}
\caption{Impact of the loss function evaluated on MMPD (in BPMs).}
\label{table:loss_ablation}
\end{minipage}%
\hfill
\begin{minipage}{0.21\textwidth}
\centering

\renewcommand{\arraystretch}{1.15}
\adjustbox{width=\textwidth}{
\begin{tabular}{ccc|c}
\hline
$CZT$ & $FFT$ & $ZOCA$ & MAE $\downarrow$ \\ \hline
✗  &  ✓ & ✗   &  14.50  \\ \hline
✓  &  ✗ & ✗   & 9.15  \\ \hline
✗ &  ✓  & ✓   & 13.07  \\ \hline
✓  &  ✗ & ✓   &  \textbf{8.85} \\ \hline
\end{tabular}
}
\caption{Ablation study of ZOCA and CZT influence.}
\label{tab:self_attention_results}
\end{minipage}
\end{table}

\noindent \textbf{Impact of ZOCA and CZT influence.} As shown in Table \ref{tab:self_attention_results}, the incorporation of the proposed ZOCA and CZT notably enhances BeatFormer's cross-dataset evaluation performance. Specifically, when comparing the ZOCA block to a standard complex attention mechanism \cite{yang2020complex}, we observe that ZOCA not only achieves superior results but does so with fewer parameters, thanks to its inherent constraints. Meanwhile, the CZT is compared with the standard FFT, denoting a notable performance improvement (approximately 3 BPM). This improvement can be linked to CZT's zoomed property, particularly with short window sizes.

\noindent \textbf{Impact of video transformations in SCL training.} Figure \ref{fig:transformations_ex} illustrates the impact of each video transformation on SCL training, both individually and in combination, across the MMPD splits and the entire dataset. While the fusion of all transformations achieves the best results for the stationary and rotation splits, some transformations prove more influential than others. Temporal flipping and the LAB color space appear to be the most impactful for SCL training, while, the spatial transformation yields the best results in the walking split, the most challenging scenario. %Lastly, although the HSV color space transformation results in the highest HR error, being less effective, its performance remains comparable to the other transformations. 

\noindent Additional ablation studies regarding the configuration of the BeatFormer can be found in the supplementary material.

\subsection{Visualization and Discussion}
\begin{figure}[t]
    \centering
\includegraphics[width=0.48\textwidth]{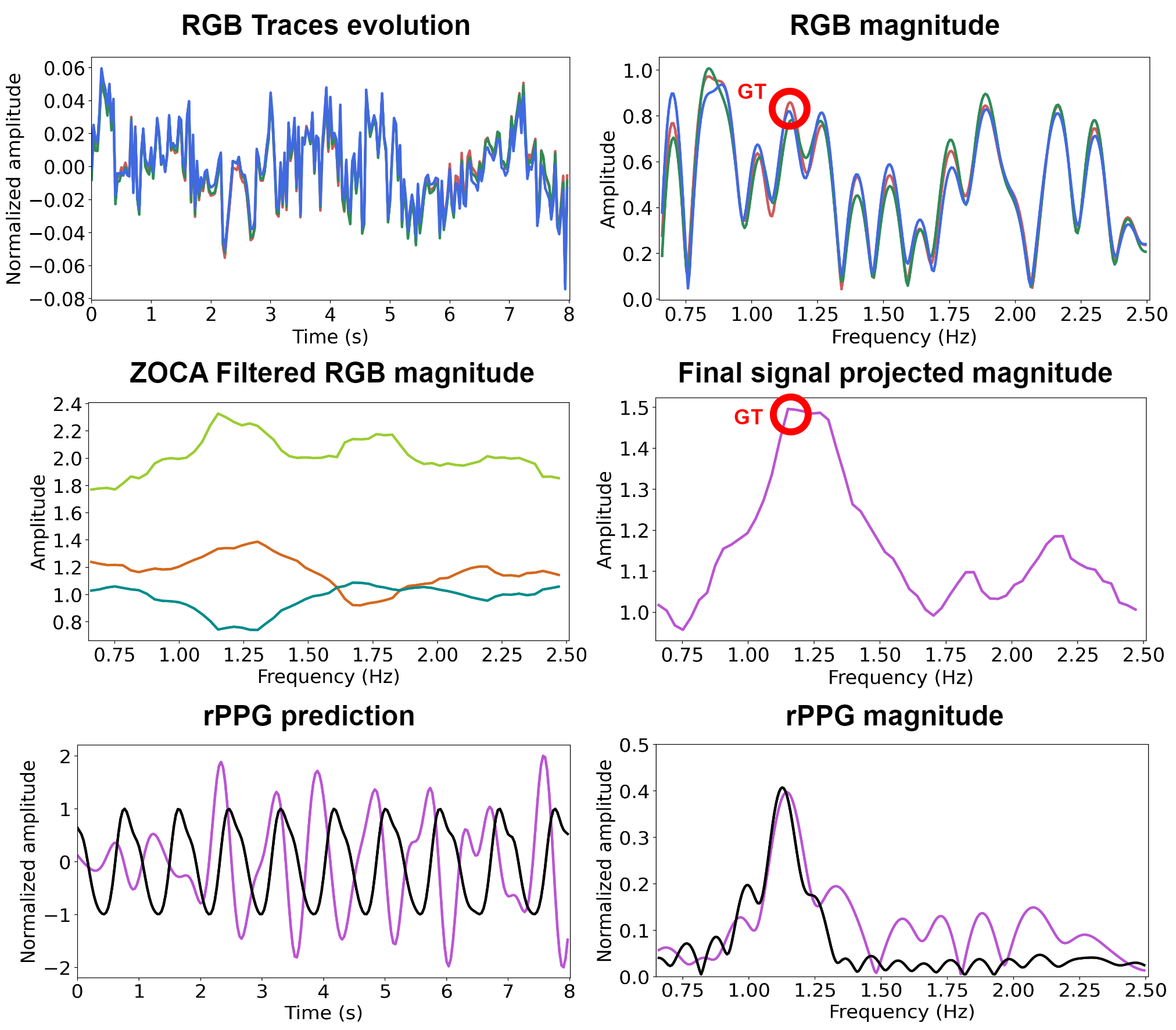}
    \caption{MMPD cross-dataset inference example. Top: RGB trace evolution and spectrum. Middle: Filtered RGB spectrum after ZOCA blocks and final signal projection. Bottom: Predicted rPPG signal (\textcolor{orchidmagenta}{\textbf{magenta}}) and PPG ground truth (\textbf{black}) comparison in time and frequency domains. Red circle denotes heart rate GT.
    } 
    \label{fig:projection}
\end{figure}
Figure \ref{fig:projection} shows a rotation sequence from MMPD processed with the unsupervised BeatFormer-SCL. On the top-right, we note that the actual pulse rate is masked due to the motion. In the middle, the resulting RGB-filtered spectrum obtained after ZOCA blocks is shown. Unlike chrominance models \cite{wang2016algorithmic, de2013robust}, which enforce a fixed RGB channel weighting (prioritizing the green channel), BeatFormer adaptively learns the green channel’s significance by emphasizing its frequency components, as seen in the filtered RGB magnitude. The final projection spectrum yields a peak at the correct pulse rate, achieving effective RGB filtering. Notably, the extracted rPPG signal remains invariant to the Pulse Transit Time (PTT), shown in the phase difference between predicted and ground truth signals. This happens because BeatFormer imposes priors to derive pulse content from the facial region rather than being biased by wrist-based PPG signals through temporal losses.

While this work introduces a novel approach, several directions remain for future exploration. We rely on a facial detector for skin area tracking, which, despite its robustness (shown in our supplementary material), adds preprocessing load that may impact efficiency. Incorporating automatic skin detection, such as temporal differentiation \cite{chen2018deepphys}, may reduce this dependency. In addition, our findings suggest that smaller window sizes benefit performance (shown in supplementary material), likely due to motion influence in the MMPD dataset. Future work could explore multi-branch architectures with different temporal resolutions to capture both short- and long-term information. Lastly, improving frequency filtering by enhancing frequency embeddings or incorporating a sparsity loss to maximize spectral power within a single frequency bin could refine signal quality.

%bin could further refine signal quality
\section{Conclusions}

We introduce BeatFormer, an efficient rPPG estimation model resilient to motion. By combining zoomed orthonormal complex attention with frequency-domain energy measurement, it integrates handcrafted priors with data-driven modeling. Our results show that constraining training with physiological priors for frequency RGB filters improves efficiency and robustness to motion bias. We also propose a novel spectral contrastive learning approach, enabling label-free training with comparable performance to labeled methods. This work highlights the benefits of frequency domain and CZT for advancing remote PPG signal recovery.

%In this work, we introduce BeatFormer, a highly efficient model for rPPG estimation resilient to motion. BeatFormer integrates zoomed orthonormal complex attention filters with frequency-domain energy measurement, while also incorporating explicit priors through spectral contrastive learning (SCL) for label-free training. This approach enhances both model efficiency and data efficiency, improving robustness to biases. Our results reveal that constraining the training through implicit and explicit priors with the representational power of data-driven models enables a lightweight yet highly effective solution while surpassing most state-of-the-art methods. Furthermore, the use of the frequency domain and CZT demonstrates promising potential for advancing remote physiological signal recovery.

%\input{sec/0_abstract}    
%\input{sec/1_intro}
%\input{sec/2_formatting}
%\input{sec/3_finalcopy}
{
    \small
    \bibliographystyle{ieeenat_fullname}
    \bibliography{References}

\begin{thebibliography}{82}
\providecommand{\natexlab}[1]{#1}
\providecommand{\url}[1]{\texttt{#1}}
\expandafter\ifx\csname urlstyle\endcsname\relax
  \providecommand{\doi}[1]{doi: #1}\else
  \providecommand{\doi}{doi: \begingroup \urlstyle{rm}\Url}\fi

\bibitem[Benezeth et~al.(2018)Benezeth, Li, Macwan, Nakamura, Gomez, and Yang]{benezeth2018remote}
Yannick Benezeth, Peixi Li, Richard Macwan, Keisuke Nakamura, Randy Gomez, and Fan Yang.
\newblock Remote heart rate variability for emotional state monitoring.
\newblock In \emph{EMBS Int. Conf. Biomed. Health Inform. BHI}, pages 153--156. IEEE, 2018.

\bibitem[Birla et~al.(2023)Birla, Shukla, Gupta, and Gupta]{birla2023alpine}
Lokendra Birla, Sneha Shukla, Anup~Kumar Gupta, and Puneet Gupta.
\newblock Alpine: Improving remote heart rate estimation using contrastive learning.
\newblock In \emph{WACV}, pages 5029--5038, 2023.

\bibitem[Bobbia et~al.(2019)Bobbia, Macwan, Benezeth, Mansouri, and Dubois]{bobbia2019unsupervised}
Serge Bobbia, Richard Macwan, Yannick Benezeth, Alamin Mansouri, and Julien Dubois.
\newblock Unsupervised skin tissue segmentation for remote photoplethysmography.
\newblock \emph{Pattern Recognit. Lett.}, 124:\penalty0 82--90, 2019.

\bibitem[Cantrill et~al.(2024)Cantrill, Ahmedt-Aristizabal, Petersson, Suominen, and Armin]{cantrill2024orientation}
Sam Cantrill, David Ahmedt-Aristizabal, Lars Petersson, Hanna Suominen, and Mohammad~Ali Armin.
\newblock Orientation-conditioned facial texture mapping for video-based facial remote photoplethysmography estimation.
\newblock In \emph{CVPR}, pages 354--363, 2024.

\bibitem[Chari et~al.(2024)Chari, Harish, Armouti, Vilesov, Sarda, Jalilian, and Kadambi]{chari2024implicit}
Pradyumna Chari, Anirudh~Bindiganavale Harish, Adnan Armouti, Alexander Vilesov, Sanjit Sarda, Laleh Jalilian, and Achuta Kadambi.
\newblock Implicit neural models to extract heart rate from video.
\newblock In \emph{ECCV}, 2024.

\bibitem[Chen et~al.(2020)Chen, Kornblith, Norouzi, and Hinton]{chen2020simple}
Ting Chen, Simon Kornblith, Mohammad Norouzi, and Geoffrey Hinton.
\newblock A simple framework for contrastive learning of visual representations.
\newblock In \emph{ICML}, pages 1597--1607. PmLR, 2020.

\bibitem[Chen and McDuff(2018)]{chen2018deepphys}
Weixuan Chen and Daniel McDuff.
\newblock Deepphys: Video-based physiological measurement using convolutional attention networks.
\newblock In \emph{ECCV}, pages 349--365, 2018.

\bibitem[Comas et~al.(2022)Comas, Ruiz, and Sukno]{comas2022efficient}
Joaquim Comas, Adria Ruiz, and Federico Sukno.
\newblock Efficient remote photoplethysmography with temporal derivative modules and time-shift invariant loss.
\newblock In \emph{CVPR}, pages 2182--2191, 2022.

\bibitem[Comas et~al.(2024{\natexlab{a}})Comas, Alomar, Ruiz, and Sukno]{comas2024physflow}
Joaquim Comas, Antonia Alomar, Adria Ruiz, and Federico Sukno.
\newblock Physflow: Skin tone transfer for remote heart rate estimation through conditional normalizing flows.
\newblock \emph{BMVC}, 2024{\natexlab{a}}.

\bibitem[Comas et~al.(2024{\natexlab{b}})Comas, Ruiz, and Sukno]{comas2024deep}
Joaquim Comas, Adria Ruiz, and Federico Sukno.
\newblock Deep pulse-signal magnification for remote heart rate estimation in compressed videos.
\newblock \emph{arXiv preprint arXiv:2405.02652}, 2024{\natexlab{b}}.

\bibitem[Comas et~al.(2024{\natexlab{c}})Comas, Ruiz, and Sukno]{comasdeepzoom2024}
Joaquim Comas, Adria Ruiz, and Federico Sukno.
\newblock Deep adaptative spectral zoom for improved remote heart rate estimation.
\newblock In \emph{FG}, 2024{\natexlab{c}}.

\bibitem[Dall’Olio et~al.(2020)Dall’Olio, Curti, Remondini, Safi~Harb, Asselbergs, Castellani, and Uh]{dall2020prediction}
Lorenzo Dall’Olio, Nico Curti, Daniel Remondini, Yosef Safi~Harb, Folkert~W Asselbergs, Gastone Castellani, and Hae-Won Uh.
\newblock Prediction of vascular aging based on smartphone acquired ppg signals.
\newblock \emph{Scientific reports}, 10:\penalty0 1--10, 2020.

\bibitem[Dasari et~al.(2021)Dasari, Prakash, Jeni, and Tucker]{dasari2021evaluation}
Ananyananda Dasari, Sakthi Kumar~Arul Prakash, L{\'a}szl{\'o}~A Jeni, and Conrad~S Tucker.
\newblock Evaluation of biases in remote photoplethysmography methods.
\newblock \emph{NPJ digital medicine}, 4\penalty0 (1):\penalty0 91, 2021.

\bibitem[De~Haan and Jeanne(2013)]{de2013robust}
Gerard De~Haan and Vincent Jeanne.
\newblock Robust pulse rate from chrominance-based rppg.
\newblock \emph{IEEE Trans. Biomed. Eng.}, 60\penalty0 (10):\penalty0 2878--2886, 2013.

\bibitem[Du et~al.(2023)Du, Liu, Zhang, and Yuen]{du2023dual}
Jingda Du, Si-Qi Liu, Bochao Zhang, and Pong~C Yuen.
\newblock Dual-bridging with adversarial noise generation for domain adaptive rppg estimation.
\newblock In \emph{CVPR}, pages 10355--10364, 2023.

\bibitem[Gideon and Stent(2021)]{gideon2021way}
John Gideon and Simon Stent.
\newblock The way to my heart is through contrastive learning: Remote photoplethysmography from unlabelled video.
\newblock In \emph{ICCV}, pages 3995--4004, 2021.

\bibitem[Gupta et~al.(2023)Gupta, Kumar, Birla, and Gupta]{gupta2023radiant}
Anup~Kumar Gupta, Rupesh Kumar, Lokendra Birla, and Puneet Gupta.
\newblock Radiant: Better rppg estimation using signal embeddings and transformer.
\newblock In \emph{WACV}, pages 4976--4986, 2023.

\bibitem[Hou et~al.(2017)Hou, Yu, and Samaras]{hou2017squared}
Le Hou, Chen-Ping Yu, and Dimitris Samaras.
\newblock Squared earth movers distance loss for training deep neural networks on ordered-classes.
\newblock In \emph{NIPS Workshop}, 2017.

\bibitem[Hsieh et~al.(2022)Hsieh, Chung, and Hsu]{hsieh2022augmentation}
Cheng-Ju Hsieh, Wei-Hao Chung, and Chiou-Ting Hsu.
\newblock Augmentation of rppg benchmark datasets: Learning to remove and embed rppg signals via double cycle consistent learning from unpaired facial videos.
\newblock In \emph{ECCV}, pages 372--387. Springer, 2022.

\bibitem[Huang et~al.(2021)Huang, Chen, Lin, Juang, Xing, Wang, and Wang]{huang2021neonatal}
Bin Huang, Weihai Chen, Chun-Liang Lin, Chia-Feng Juang, Yuanping Xing, Yanting Wang, and Jianhua Wang.
\newblock A neonatal dataset and benchmark for non-contact neonatal heart rate monitoring based on spatio-temporal neural networks.
\newblock \emph{Engineering Applications of Artificial Intelligence}, 106:\penalty0 104447, 2021.

\bibitem[Kang et~al.()Kang, Lee, Kim, Chung, and Yoon]{kangintroducing}
Bong~Gyun Kang, Dongjun Lee, HyunGi Kim, Dohyun Chung, and Sungroh Yoon.
\newblock Introducing spectral attention for long-range dependency in time series forecasting.
\newblock In \emph{NeurIPS}.

\bibitem[Lee et~al.(2020)Lee, Chen, and Lee]{lee2020meta}
Eugene Lee, Evan Chen, and Chen-Yi Lee.
\newblock Meta-rppg: Remote heart rate estimation using a transductive meta-learner.
\newblock In \emph{ECCV}, pages 392--409. Springer, 2020.

\bibitem[Lee-Thorp et~al.(2021)Lee-Thorp, Ainslie, Eckstein, and Ontanon]{lee2021fnet}
James Lee-Thorp, Joshua Ainslie, Ilya Eckstein, and Santiago Ontanon.
\newblock Fnet: Mixing tokens with fourier transforms.
\newblock \emph{arXiv preprint arXiv:2105.03824}, 2021.

\bibitem[Li et~al.(2023)Li, Yu, and Shi]{li2023learning}
Jianwei Li, Zitong Yu, and Jingang Shi.
\newblock Learning motion-robust remote photoplethysmography through arbitrary resolution videos.
\newblock In \emph{AAAI}, pages 1334--1342, 2023.

\bibitem[Li et~al.(2014)Li, Chen, Zhao, and Pietikainen]{li2014remote}
Xiaobai Li, Jie Chen, Guoying Zhao, and Matti Pietikainen.
\newblock Remote heart rate measurement from face videos under realistic situations.
\newblock In \emph{CVPR}, pages 4264--4271, 2014.

\bibitem[Liu et~al.(2022)Liu, Yang, Fu, and Qian]{liu2022learning}
Chengxu Liu, Huan Yang, Jianlong Fu, and Xueming Qian.
\newblock Learning trajectory-aware transformer for video super-resolution.
\newblock In \emph{CVPR}, pages 5687--5696, 2022.

\bibitem[Liu et~al.(2025)Liu, Tang, Chen, Li, Qi, Li, Wang, Gan, Wang, and Chen]{liu2025spiking}
Mingxuan Liu, Jiankai Tang, Yongli Chen, Haoxiang Li, Jiahao Qi, Siwei Li, Kegang Wang, Jie Gan, Yuntao Wang, and Hong Chen.
\newblock Spiking-physformer: Camera-based remote photoplethysmography with parallel spike-driven transformer.
\newblock \emph{Neural Networks}, page 107128, 2025.

\bibitem[Liu et~al.(2021{\natexlab{a}})Liu, Deng, Huang, Shi, Lu, Sun, Wang, Dai, and Li]{liu2021fuseformer}
Rui Liu, Hanming Deng, Yangyi Huang, Xiaoyu Shi, Lewei Lu, Wenxiu Sun, Xiaogang Wang, Jifeng Dai, and Hongsheng Li.
\newblock Fuseformer: Fusing fine-grained information in transformers for video inpainting.
\newblock In \emph{ICCV}, pages 14040--14049, 2021{\natexlab{a}}.

\bibitem[Liu et~al.(2020)Liu, Fromm, Patel, and McDuff]{liu2020multi}
Xin Liu, Josh Fromm, Shwetak Patel, and Daniel McDuff.
\newblock Multi-task temporal shift attention networks for on-device contactless vitals measurement.
\newblock \emph{NeurIPS}, 33:\penalty0 19400--19411, 2020.

\bibitem[Liu et~al.(2021{\natexlab{b}})Liu, Jiang, Fromm, Xu, Patel, and McDuff]{liu2021metaphys}
Xin Liu, Ziheng Jiang, Josh Fromm, Xuhai Xu, Shwetak Patel, and Daniel McDuff.
\newblock Metaphys: few-shot adaptation for non-contact physiological measurement.
\newblock In \emph{CHIL}, pages 154--163, 2021{\natexlab{b}}.

\bibitem[Liu et~al.(2023{\natexlab{a}})Liu, Hill, Jiang, Patel, and McDuff]{liu2023efficientphys}
Xin Liu, Brian Hill, Ziheng Jiang, Shwetak Patel, and Daniel McDuff.
\newblock Efficientphys: Enabling simple, fast and accurate camera-based cardiac measurement.
\newblock In \emph{WACV}, pages 5008--5017, 2023{\natexlab{a}}.

\bibitem[Liu et~al.(2023{\natexlab{b}})Liu, Narayanswamy, Paruchuri, Zhang, Tang, Zhang, Sengupta, Patel, Wang, and McDuff]{liu2023rppg}
Xin Liu, Girish Narayanswamy, Akshay Paruchuri, Xiaoyu Zhang, Jiankai Tang, Yuzhe Zhang, Roni Sengupta, Shwetak Patel, Yuntao Wang, and Daniel McDuff.
\newblock rppg-toolbox: Deep remote ppg toolbox.
\newblock \emph{NeurIPS}, 36:\penalty0 68485--68510, 2023{\natexlab{b}}.

\bibitem[Liu et~al.(2024)Liu, Zhang, Yu, Lu, Yue, and Yang]{liu2024rppg}
Xin Liu, Yuting Zhang, Zitong Yu, Hao Lu, Huanjing Yue, and Jingyu Yang.
\newblock rppg-mae: Self-supervised pretraining with masked autoencoders for remote physiological measurements.
\newblock \emph{IEEE Transactions on Multimedia}, 2024.

\bibitem[Loshchilov and Hutter(2022)]{loshchilov2022sgdr}
Ilya Loshchilov and Frank Hutter.
\newblock Sgdr: Stochastic gradient descent with warm restarts.
\newblock In \emph{ICLR}, 2022.

\bibitem[Lu et~al.(2021)Lu, Han, and Zhou]{lu2021dual}
Hao Lu, Hu Han, and S~Kevin Zhou.
\newblock Dual-gan: Joint bvp and noise modeling for remote physiological measurement.
\newblock In \emph{CVPR}, pages 12404--12413, 2021.

\bibitem[Lu et~al.(2023)Lu, Yu, Niu, and Chen]{lu2023neuron}
Hao Lu, Zitong Yu, Xuesong Niu, and Ying-Cong Chen.
\newblock Neuron structure modeling for generalizable remote physiological measurement.
\newblock In \emph{CVPR}, pages 18589--18599, 2023.

\bibitem[Lugaresi et~al.(2019)Lugaresi, Tang, Nash, McClanahan, Uboweja, Hays, Zhang, Chang, Yong, Lee, et~al.]{lugaresi2019mediapipe}
Camillo Lugaresi, Jiuqiang Tang, Hadon Nash, Chris McClanahan, Esha Uboweja, Michael Hays, Fan Zhang, Chuo-Ling Chang, Ming~Guang Yong, Juhyun Lee, et~al.
\newblock Mediapipe: A framework for building perception pipelines.
\newblock \emph{arXiv preprint arXiv:1906.08172}, 2019.

\bibitem[Luo et~al.(2024)Luo, Xie, and Yu]{luo2024physmamba}
Chaoqi Luo, Yiping Xie, and Zitong Yu.
\newblock Physmamba: Efficient remote physiological measurement with slowfast temporal difference mamba.
\newblock \emph{arXiv preprint arXiv:2409.12031}, 2024.

\bibitem[Maity et~al.(2022)Maity, Wang, Sabharwal, and Nayar]{maity2022robustppg}
Akash~Kumar Maity, Jian Wang, Ashutosh Sabharwal, and Shree~K Nayar.
\newblock Robustppg: camera-based robust heart rate estimation using motion cancellation.
\newblock \emph{Biomedical Optics Express}, 13\penalty0 (10):\penalty0 5447--5467, 2022.

\bibitem[Niu et~al.(2019)Niu, Shan, Han, and Chen]{niu2019rhythmnet}
Xuesong Niu, Shiguang Shan, Hu Han, and Xilin Chen.
\newblock Rhythmnet: End-to-end heart rate estimation from face via spatial-temporal representation.
\newblock \emph{IEEE TIP}, 29:\penalty0 2409--2423, 2019.

\bibitem[Nowara et~al.(2020)Nowara, Marks, Mansour, and Veeraraghavan]{nowara2020near}
Ewa~M Nowara, Tim~K Marks, Hassan Mansour, and Ashok Veeraraghavan.
\newblock Near-infrared imaging photoplethysmography during driving.
\newblock \emph{IEEE transactions on intelligent transportation systems}, 23\penalty0 (4):\penalty0 3589--3600, 2020.

\bibitem[Paruchuri et~al.(2024)Paruchuri, Liu, Pan, Patel, McDuff, and Sengupta]{paruchuri2024motion}
Akshay Paruchuri, Xin Liu, Yulu Pan, Shwetak Patel, Daniel McDuff, and Soumyadip Sengupta.
\newblock Motion matters: Neural motion transfer for better camera physiological measurement.
\newblock In \emph{WACV}, pages 5933--5942, 2024.

\bibitem[Paszke et~al.(2019)Paszke, Gross, Massa, Lerer, Bradbury, Chanan, Killeen, Lin, Gimelshein, Antiga, et~al.]{paszke2019pytorch}
Adam Paszke, Sam Gross, Francisco Massa, Adam Lerer, James Bradbury, Gregory Chanan, Trevor Killeen, Zeming Lin, Natalia Gimelshein, Luca Antiga, et~al.
\newblock Pytorch: An imperative style, high-performance deep learning library.
\newblock \emph{NeurIPS}, 32, 2019.

\bibitem[Perepelkina et~al.(2020)Perepelkina, Artemyev, Churikova, and Grinenko]{perepelkina2020hearttrack}
Olga Perepelkina, Mikhail Artemyev, Marina Churikova, and Mikhail Grinenko.
\newblock Hearttrack: Convolutional neural network for remote video-based heart rate monitoring.
\newblock In \emph{CVPRW}, pages 288--289, 2020.

\bibitem[Plizzari et~al.(2021)Plizzari, Cannici, and Matteucci]{plizzari2021spatial}
Chiara Plizzari, Marco Cannici, and Matteo Matteucci.
\newblock Spatial temporal transformer network for skeleton-based action recognition.
\newblock In \emph{ICPR}, pages 694--701. Springer, 2021.

\bibitem[Poh et~al.(2010{\natexlab{a}})Poh, McDuff, and Picard]{poh2010advancements}
Ming-Zher Poh, Daniel~J McDuff, and Rosalind~W Picard.
\newblock Advancements in noncontact, multiparameter physiological measurements using a webcam.
\newblock \emph{IEEE Trans. Biomed. Eng.}, 58\penalty0 (1):\penalty0 7--11, 2010{\natexlab{a}}.

\bibitem[Poh et~al.(2010{\natexlab{b}})Poh, McDuff, and Picard]{poh2010non}
Ming-Zher Poh, Daniel~J McDuff, and Rosalind~W Picard.
\newblock Non-contact, automated cardiac pulse measurements using video imaging and blind source separation.
\newblock \emph{Optics express}, 18\penalty0 (10):\penalty0 10762--10774, 2010{\natexlab{b}}.

\bibitem[Qian et~al.(2021)Qian, Meng, Gong, Yang, Wang, Belongie, and Cui]{qian2021spatiotemporal}
Rui Qian, Tianjian Meng, Boqing Gong, Ming-Hsuan Yang, Huisheng Wang, Serge Belongie, and Yin Cui.
\newblock Spatiotemporal contrastive video representation learning.
\newblock In \emph{CVPR}, pages 6964--6974, 2021.

\bibitem[Qin et~al.(2021)Qin, Zhang, Wu, and Li]{qin2021fcanet}
Zequn Qin, Pengyi Zhang, Fei Wu, and Xi Li.
\newblock Fcanet: Frequency channel attention networks.
\newblock In \emph{ICCV}, pages 783--792, 2021.

\bibitem[Rabiner et~al.(1969)Rabiner, Schafer, and Rader]{rabiner1969chirp}
L Rabiner, R~W Schafer, and C Rader.
\newblock The chirp z-transform algorithm.
\newblock \emph{IEEE transactions on audio and electroacoustics}, 17\penalty0 (2):\penalty0 86--92, 1969.

\bibitem[Savic and Zhao(2024)]{savic2024rs}
Marko Savic and Guoying Zhao.
\newblock Rs-rppg: robust self-supervised learning for rppg.
\newblock In \emph{2024 IEEE 18th International Conference on Automatic Face and Gesture Recognition (FG)}, pages 1--10. IEEE, 2024.

\bibitem[Schroff et~al.(2015)Schroff, Kalenichenko, and Philbin]{schroff2015facenet}
Florian Schroff, Dmitry Kalenichenko, and James Philbin.
\newblock Facenet: A unified embedding for face recognition and clustering.
\newblock In \emph{CVPR}, pages 815--823, 2015.

\bibitem[Song et~al.(2021)Song, Chen, Cheng, Li, Liu, and Chen]{song2021pulsegan}
Rencheng Song, Huan Chen, Juan Cheng, Chang Li, Yu Liu, and Xun Chen.
\newblock Pulsegan: Learning to generate realistic pulse waveforms in remote photoplethysmography.
\newblock \emph{IEEE J.Biomed.Health Inform.}, 25\penalty0 (5):\penalty0 1373--1384, 2021.

\bibitem[{\v{S}}petl{\'\i}k et~al.(2018){\v{S}}petl{\'\i}k, Franc, and Matas]{vspetlik2018visual}
Radim {\v{S}}petl{\'\i}k, Vojtech Franc, and Jir{\'\i} Matas.
\newblock Visual heart rate estimation with convolutional neural network.
\newblock In \emph{BMVC}, 2018.

\bibitem[Stricker et~al.(2014)Stricker, M{\"u}ller, and Gross]{stricker2014non}
Ronny Stricker, Steffen M{\"u}ller, and Horst-Michael Gross.
\newblock Non-contact video-based pulse rate measurement on a mobile service robot.
\newblock In \emph{RO-MAN}, pages 1056--1062. IEEE, 2014.

\bibitem[Sun and Li(2022)]{sun2022contrast}
Zhaodong Sun and Xiaobai Li.
\newblock Contrast-phys: Unsupervised video-based remote physiological measurement via spatiotemporal contrast.
\newblock In \emph{ECCV}, pages 492--510. Springer, 2022.

\bibitem[Sun and Li(2024)]{sun2024contrast}
Zhaodong Sun and Xiaobai Li.
\newblock Contrast-phys+: Unsupervised and weakly-supervised video-based remote physiological measurement via spatiotemporal contrast.
\newblock \emph{IEEE Transactions on Pattern Analysis and Machine Intelligence}, 2024.

\bibitem[Takano and Ohta(2007)]{takano2007heart}
Chihiro Takano and Yuji Ohta.
\newblock Heart rate measurement based on a time-lapse image.
\newblock \emph{Medical engineering \& physics}, 29\penalty0 (8):\penalty0 853--857, 2007.

\bibitem[Tan et~al.(2022)Tan, Xie, Jiang, and Zhou]{tan2022real}
Zhi-Hao Tan, Yi Xie, Yuan Jiang, and Zhi-Hua Zhou.
\newblock Real-valued backpropagation is unsuitable for complex-valued neural networks.
\newblock \emph{NeurIPS}, 35:\penalty0 34052--34063, 2022.

\bibitem[Tang et~al.(2023)Tang, Chen, Wang, Shi, Patel, McDuff, and Liu]{tang2023mmpd}
Jiankai Tang, Kequan Chen, Yuntao Wang, Yuanchun Shi, Shwetak Patel, Daniel McDuff, and Xin Liu.
\newblock Mmpd: Multi-domain mobile video physiology dataset.
\newblock \emph{arXiv preprint arXiv:2302.03840}, 2023.

\bibitem[Tarvainen et~al.(2002)Tarvainen, Ranta-Aho, and Karjalainen]{tarvainen2002advanced}
Mika~P Tarvainen, Perttu~O Ranta-Aho, and Pasi~A Karjalainen.
\newblock An advanced detrending method with application to hrv analysis.
\newblock \emph{IEEE transactions on biomedical engineering}, 49\penalty0 (2):\penalty0 172--175, 2002.

\bibitem[Trabelsi et~al.(2018)Trabelsi, Bilaniuk, Zhang, Serdyuk, Subramanian, Santos, Mehri, Rostamzadeh, Bengio, and Pal]{trabelsi2018deep}
Chiheb Trabelsi, Olexa Bilaniuk, Ying Zhang, Dmitriy Serdyuk, Sandeep Subramanian, Joao~Felipe Santos, Soroush Mehri, Negar Rostamzadeh, Yoshua Bengio, and Christopher~J Pal.
\newblock Deep complex networks.
\newblock In \emph{ICLR}, 2018.

\bibitem[Tsouri and Li(2015)]{tsouri2015benefits}
Gill~R Tsouri and Zheng Li.
\newblock On the benefits of alternative color spaces for noncontact heart rate measurements using standard red-green-blue cameras.
\newblock \emph{Journal of biomedical optics}, 20\penalty0 (4):\penalty0 048002--048002, 2015.

\bibitem[Vaswani(2017)]{vaswani2017attention}
A Vaswani.
\newblock Attention is all you need.
\newblock \emph{NeurIPS}, 2017.

\bibitem[Verkruysse et~al.(2008)Verkruysse, Svaasand, and Nelson]{verkruysse2008remote}
Wim Verkruysse, Lars~O Svaasand, and J~Stuart Nelson.
\newblock Remote plethysmographic imaging using ambient light.
\newblock \emph{Optics express}, 16\penalty0 (26):\penalty0 21434--21445, 2008.

\bibitem[Wang et~al.(2023)Wang, Sun, Hao, Pan, and Jia]{wang2023transphys}
Rui-Xuan Wang, Hong-Mei Sun, Rong-Rong Hao, Ang Pan, and Rui-Sheng Jia.
\newblock Transphys: Transformer-based unsupervised contrastive learning for remote heart rate measurement.
\newblock \emph{Biomedical Signal Processing and Control}, 86:\penalty0 105058, 2023.

\bibitem[Wang et~al.(2016)Wang, den Brinker, Stuijk, and De~Haan]{wang2016algorithmic}
Wenjin Wang, Albertus~C den Brinker, Sander Stuijk, and Gerard De~Haan.
\newblock Algorithmic principles of remote ppg.
\newblock \emph{IEEE Trans. Biomed. Eng.}, 64\penalty0 (7):\penalty0 1479--1491, 2016.

\bibitem[Wang et~al.(2017{\natexlab{a}})Wang, den Brinker, Stuijk, and de~Haan]{wang2017color}
Wenjin Wang, Albertus~C den Brinker, Sander Stuijk, and Gerard de Haan.
\newblock Color-distortion filtering for remote photoplethysmography.
\newblock In \emph{FG}, pages 71--78. IEEE, 2017{\natexlab{a}}.

\bibitem[Wang et~al.(2017{\natexlab{b}})Wang, den Brinker, Stuijk, and de~Haan]{wang2017robust}
Wenjin Wang, Albertus~C den Brinker, Sander Stuijk, and Gerard de Haan.
\newblock Robust heart rate from fitness videos.
\newblock \emph{Physiological measurement}, 38\penalty0 (6):\penalty0 1023, 2017{\natexlab{b}}.

\bibitem[Wolter and Yao(2018)]{wolter2018complex}
Moritz Wolter and Angela Yao.
\newblock Complex gated recurrent neural networks.
\newblock \emph{NeurIPS}, 31, 2018.

\bibitem[Wu et~al.(2023)Wu, Chiu, Wu, Lin, Ho, Chung, and Wu]{wu2023motion}
Yi-Chiao Wu, Li-Wen Chiu, Bing-Fei Wu, Linda Li-Chuan Lin, Tsai-Hsuan Ho, Meng-Liang Chung, and Shou-Fang Wu.
\newblock Motion robust remote photoplethysmography measurement during exercise for contactless physical activity intensity detection.
\newblock \emph{IEEE TIM}, 72:\penalty0 1--14, 2023.

\bibitem[Yang et~al.(2020)Yang, Ma, Li, Tsai, and Salakhutdinov]{yang2020complex}
Muqiao Yang, Martin~Q Ma, Dongyu Li, Yao-Hung~Hubert Tsai, and Ruslan Salakhutdinov.
\newblock Complex transformer: A framework for modeling complex-valued sequence.
\newblock In \emph{ICASSP}, pages 4232--4236. IEEE, 2020.

\bibitem[Yang et~al.(2016)Yang, Liu, Yu, Shao, Tsow, and Tao]{yang2016motion}
Yuting Yang, Chenbin Liu, Hui Yu, Dangdang Shao, Francis Tsow, and Nongjian Tao.
\newblock Motion robust remote photoplethysmography in cielab color space.
\newblock \emph{Journal of biomedical optics}, 21\penalty0 (11):\penalty0 117001--117001, 2016.

\bibitem[Yu et~al.(2019{\natexlab{a}})Yu, Li, and Zhao]{Yu2019RemotePS}
Z. Yu, Xiao-Bai Li, and G. Zhao.
\newblock Remote photoplethysmograph signal measurement from facial videos using spatio-temporal networks.
\newblock In \emph{BMVC}, 2019{\natexlab{a}}.

\bibitem[Yu et~al.(2019{\natexlab{b}})Yu, Peng, Li, Hong, and Zhao]{yu2019remote}
Zitong Yu, Wei Peng, Xiaobai Li, Xiaopeng Hong, and Guoying Zhao.
\newblock Remote heart rate measurement from highly compressed facial videos: an end-to-end deep learning solution with video enhancement.
\newblock In \emph{ICCV}, pages 151--160, 2019{\natexlab{b}}.

\bibitem[Yu et~al.(2021)Yu, Li, Wang, and Zhao]{yu2021transrppg}
Zitong Yu, Xiaobai Li, Pichao Wang, and Guoying Zhao.
\newblock Transrppg: Remote photoplethysmography transformer for 3d mask face presentation attack detection.
\newblock \emph{IEEE Signal Processing Letters}, 28:\penalty0 1290--1294, 2021.

\bibitem[Yu et~al.(2022)Yu, Shen, Shi, Zhao, Torr, and Zhao]{yu2022physformer}
Zitong Yu, Yuming Shen, Jingang Shi, Hengshuang Zhao, Philip~HS Torr, and Guoying Zhao.
\newblock Physformer: facial video-based physiological measurement with temporal difference transformer.
\newblock In \emph{CVPR}, pages 4186--4196, 2022.

\bibitem[Yu et~al.(2023)Yu, Shen, Shi, Zhao, Cui, Zhang, Torr, and Zhao]{yu2023physformer++}
Zitong Yu, Yuming Shen, Jingang Shi, Hengshuang Zhao, Yawen Cui, Jiehua Zhang, Philip Torr, and Guoying Zhao.
\newblock Physformer++: Facial video-based physiological measurement with slowfast temporal difference transformer.
\newblock \emph{IJCV}, 131\penalty0 (6):\penalty0 1307--1330, 2023.

\bibitem[Zhang et~al.(2023)Zhang, Xia, Liu, and Feng]{zhang2023demodulation}
Xiaobiao Zhang, Zhaoqiang Xia, Lili Liu, and Xiaoyi Feng.
\newblock Demodulation based transformer for rppg generation and heart rate estimation.
\newblock \emph{IEEE Signal Processing Letters}, 2023.

\bibitem[Zhao et~al.(2024)Zhao, Sun, Tian, Yang, Tao, Cheng, and Chen]{zhao2024toward}
Pengfei Zhao, Qigong Sun, Xiaolin Tian, Yige Yang, Shuo Tao, Jie Cheng, and Jiantong Chen.
\newblock Toward motion robustness: A masked attention regularization framework in remote photoplethysmography.
\newblock In \emph{CVPR}, pages 7829--7838, 2024.

\bibitem[Zhou et~al.(2020)Zhou, Krause, Blocher, and Stork]{zhou2020enhancing}
Kai Zhou, Simon Krause, Timon Blocher, and Wilhelm Stork.
\newblock Enhancing remote-ppg pulse extraction in disturbance scenarios utilizing spectral characteristics.
\newblock In \emph{CVPR}, pages 280--281, 2020.

\bibitem[Zou et~al.(2024)Zou, Guo, Chen, and Ma]{zou2024rhythmformer}
Bochao Zou, Zizheng Guo, Jiansheng Chen, and Huimin Ma.
\newblock Rhythmformer: Extracting rppg signals based on hierarchical temporal periodic transformer.
\newblock \emph{arXiv preprint arXiv:2402.12788}, 2024.

\end{thebibliography}
}

\end{document}